\documentclass[letterpaper]{article} 
\usepackage{aaai24}  
\usepackage{times}  
\usepackage{helvet}  
\usepackage{courier}  
\usepackage[hyphens]{url}  
\usepackage{graphicx} 
\usepackage{natbib}  
\usepackage{caption} 
\usepackage{algorithm}
\usepackage{algorithmic}

\usepackage{newfloat}
\usepackage{listings}
\DeclareCaptionStyle{ruled}{labelfont=normalfont,labelsep=colon,strut=off} 
\floatstyle{ruled}
\newfloat{listing}{tb}{lst}{}
\floatname{listing}{Listing}
\usepackage{mathrsfs}
\usepackage{amsmath}
\usepackage{amsfonts}
\usepackage{subfig} 
\usepackage{multirow}

\title{Region-aware Exposure Consistency Network for Mixed Exposure Correction}
\author {
    Jin Liu\textsuperscript{\rm 1},
    Huiyuan Fu\textsuperscript{\rm 1}\thanks{Corresponding author.},
    Chuanming Wang\textsuperscript{\rm 1},
    Huadong Ma\textsuperscript{\rm 1}
}
\affiliations {
\textsuperscript{\rm 1}Beijing University of Posts and Telecommunications, China \\
\{ljin,fhy,wcm,mhd\}@bupt.edu.cn
}

\usepackage{bibentry}

\begin{document}

\maketitle


\begin{abstract}
Exposure correction aims to enhance images suffering from improper exposure to achieve satisfactory visual effects.
Despite recent progress, existing methods generally mitigate either overexposure or underexposure in input images, and they still struggle to handle images with mixed exposure,
\emph{i.e.}, one image incorporates both overexposed and underexposed regions.
The mixed exposure distribution is non-uniform and leads to varying representation, which makes it challenging to address in a unified process.
In this paper, we introduce an effective Region-aware Exposure Correction Network (RECNet) that can handle mixed exposure by adaptively learning and bridging different regional exposure representations.
Specifically, to address the challenge posed by mixed exposure disparities, we develop a region-aware de-exposure module that effectively translates regional features of mixed exposure scenarios into an exposure-invariant feature space.
Simultaneously, as de-exposure operation inevitably reduces discriminative information, we introduce a mixed-scale restoration unit that integrates exposure-invariant features and unprocessed features to recover local information.
To further achieve a uniform exposure distribution in the global image, we propose an exposure contrastive regularization strategy under the constraints of intra-regional exposure consistency and inter-regional exposure continuity. 
Extensive experiments are conducted on various datasets, and the experimental results demonstrate the superiority and generalization of our proposed method.
The code is released at: https://github.com/kravrolens/RECNet.
\end{abstract}


\section{Introduction}

\begin{figure}[!t]
\centering
\subfloat[Under-/Over- Exp examples]{\includegraphics[width=0.49\columnwidth]{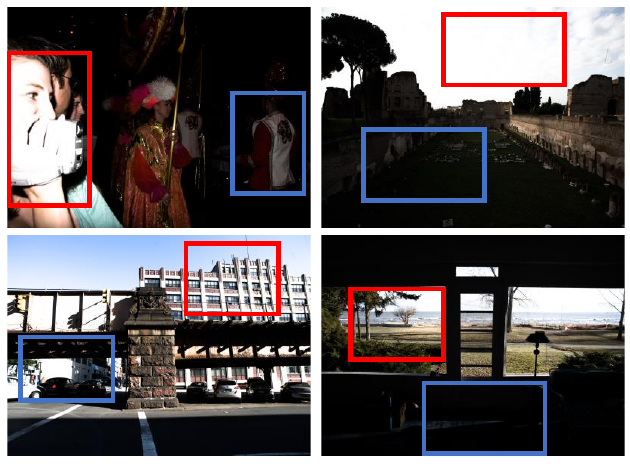}}
\subfloat[Brightness Mapping]{\includegraphics[width=0.49\columnwidth]{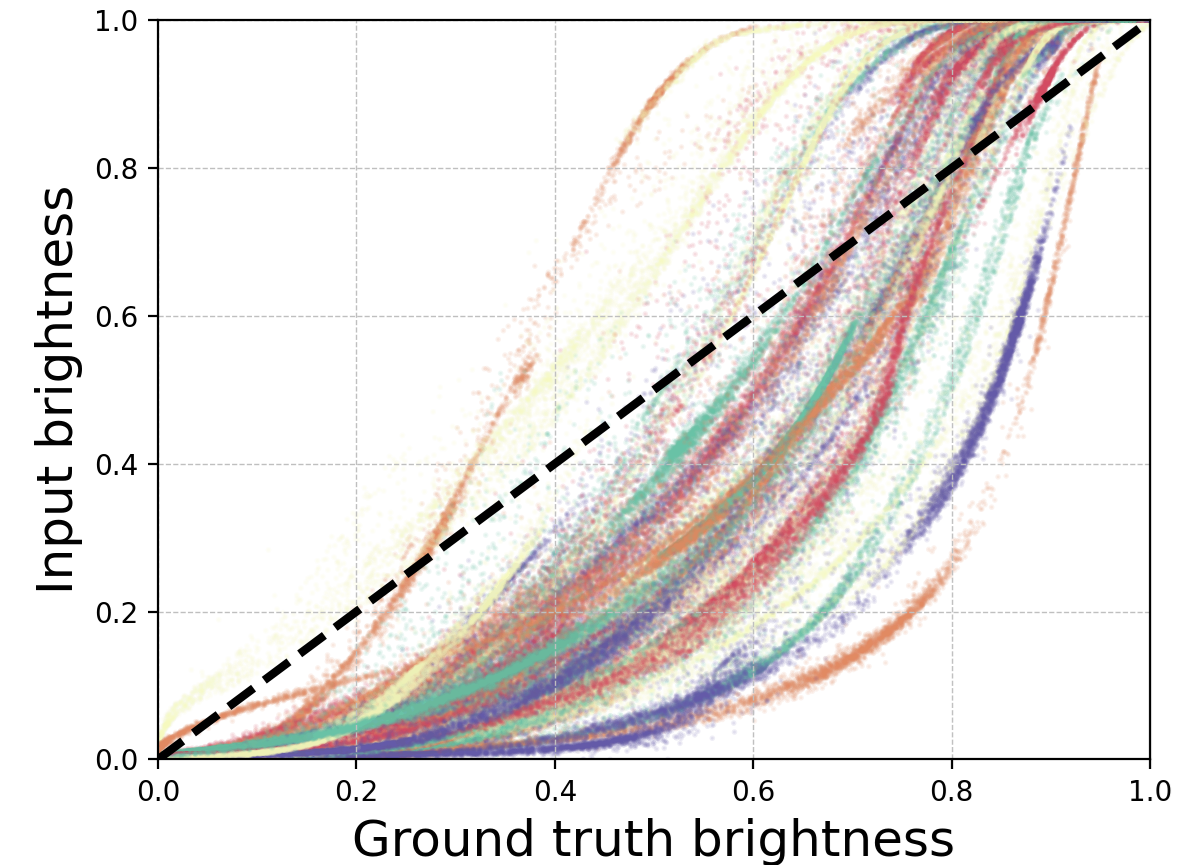}} \\
\subfloat[LCDPNet]{\includegraphics[width=0.49\columnwidth]{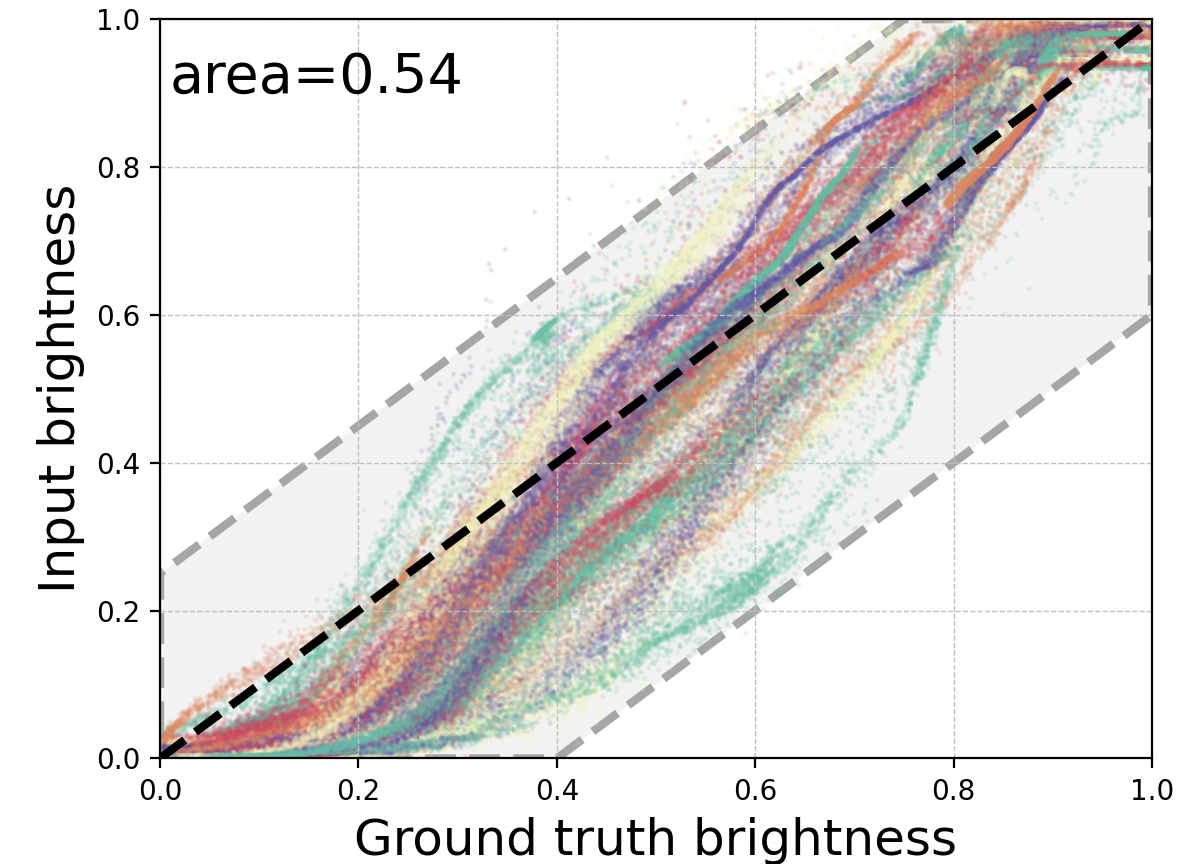}}
\subfloat[RECNet (Ours)]{\includegraphics[width=0.49\columnwidth]{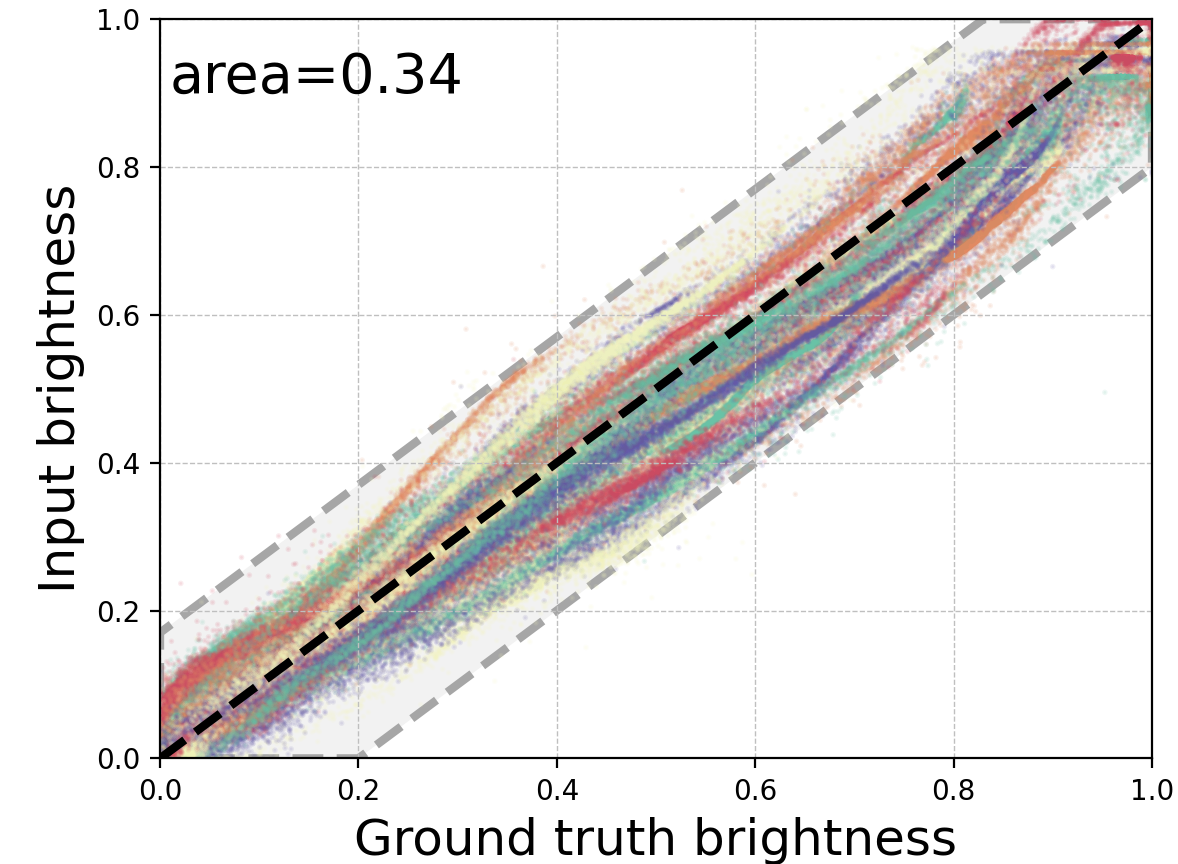}}
\caption{For a single image above in (a), it contains both overexposure (marked with red box) and underexposure (marked with blue box) regions. (b) is the input-ground truth brightness mapping curve statistics of the LCDP~\cite{wang2022local} dataset. (c) and (d) are the generated result-ground truth mapping curves of LCDPNet and our method, respectively. A smaller area represents better results with smaller correction errors.}
\label{fig:motivation}
\end{figure}

In recent years, taking photos has become universal with the rapid development of camera devices. However, images captured in unbalanced ambient lighting conditions often suffer from overexposure and underexposure problems, resulting in significant structure distortion and quality degradation. Photography experts can alleviate this problem by tuning the equipment, but it is too professional and difficult to achieve in many cases. With the development of deep learning, how to solve this problem intelligently and automatically gradually attracts the attention of researchers.

To solve this challenging but practical problem, exposure correction techniques~\cite{wei2018deep,wang2019underexposed,yang2020fidelity,wu2022uretinex} have been developed to correct varying levels of underexposed and overexposed images to normal exposure. 
However, conventional methods only emphasize either underexposed or overexposed scenes, resulting in poor generalization to the opposite exposures, limiting their deployment in practical applications.
Some methods~\cite{afifi2021learning,nsampi2021learning,huang2022deep} propose a unified network to handle both underexposure and overexposure problems. They are based on the assumption that the global illumination of the target scene is uniform, \emph{i.e.}, the image is overexposed or underexposed. However, as shown in Figure~\ref{fig:motivation}(a) and (b), both overexposure and underexposure (termed as mixed exposure) can occur in a single image when encountering unevenly lighting scenes. Due to the variations of representations in mixed exposed images, the correction procedures differ greatly from each other. Facing images with mixed exposure, these methods tend to darken the overexposed images or brighten the adversely underexposed images globally.

Despite the importance of correcting mixed exposure, this task remains relatively unexplored. The most relevant work is LCDNet~\cite{wang2022local}, which is based on the Retinex theory~\cite{land1977retinex} and uses local color distribution prior to learning the representations of overexposed and underexposed regions.
Then, it employs these learned representations to individually enhance the image for each type of exposure, and subsequently merges the outputs of different exposures to form the final result.
This method follows the pattern of localized statistics applied globally, but the degrees of underexposure and overexposure may not be the same in the image, and a post-processing fusion operation is difficult to balance them properly.
As depicted in Figure~\ref{fig:motivation}(c), even after being corrected by LCDPNet, certain regions still exhibit pronounced issues of both overexposure and underexposure.

To tackle this challenging task, we introduce Region-aware Exposure Correction Network (named RECNet) to effectively enhance various images including single or hybrid exposure conditions.
Specifically, to reduce the disparities among different exposure regions, we introduce the Region-aware De-exposure Module (RDM) to effectively map the regional features with different exposure scenarios into an exposure-invariant feature space.
It initially segments the image into distinct overexposed and underexposed regions via an exposure mask predictor and then adopts the multi-branch mask-aware instance normalization to finely align these respective regions.
Since normalization may lead to the loss of discriminative information~\cite{ren2023crossing}, we introduce the Mixed-scale Restoration Unit (MRU) that integrates the de-exposed and unprocessed features through multi-scale depth-wise convolution and channel-wise self-attention pathways, obtaining satisfactory contrast and details.
In addition, we develop the Exposure Contrastive Regularization (ECR) strategy to guarantee a uniform exposure distribution across regions. We ensure the self-exposure consistency within the regions by pulling closer the distance between the corresponding regions of the generated image and the ground-truth image in the feature space, simultaneously modeling and comparing the exposure distance between regions to ensure the exposure continuity, and finally achieving a balanced improvement for mixed exposure images.

Benefiting from the carefully designed modules and effective strategy mentioned above, our RECNet achieves superior performance on several widely used datasets. The main contributions of this paper can be summarized as follows:
\begin{itemize}
\item We propose an effective exposure correction network to improve the non-uniform exposure images that simultaneously enhances underexposed regions and suppresses overexposed regions.
\item To account for mixed exposure discrepancy, we design the region-aware de-exposure module to map regional features to exposure-invariant space, and the mixed-scale restoration unit for better information completeness.
\item We introduce an exposure contrastive regularization strategy to model the relationships across different exposed regions to achieve uniform exposure distribution.
\end{itemize}


\begin{figure*}[!t]
\centering
\includegraphics[width=\textwidth]{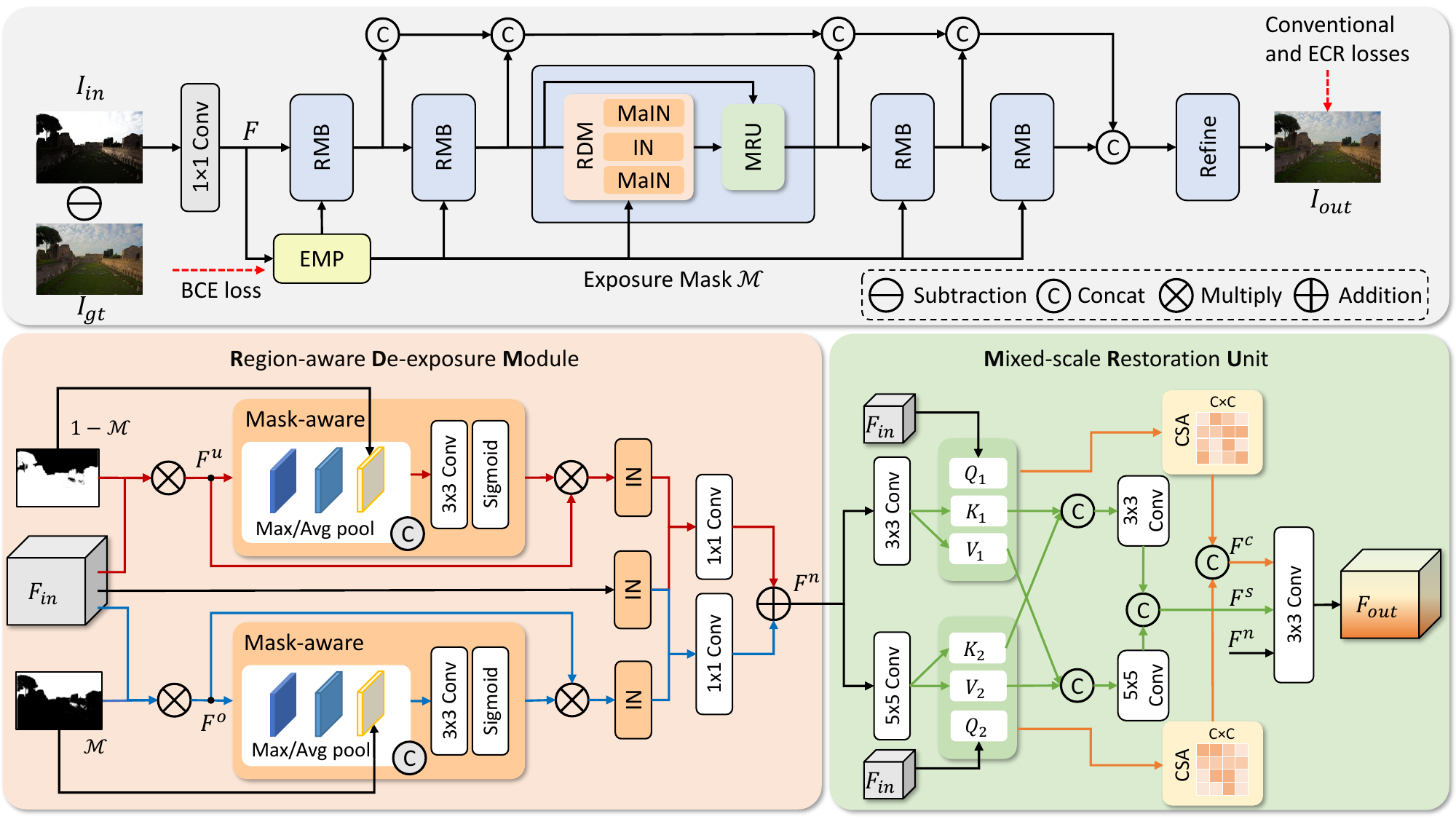}
\caption{The overall architecture of the proposed region-aware exposure correction network, which mainly contains a series of \textbf{B}locks (RMB) with \textbf{R}egion-aware De-exposure Module (RDM) and \textbf{M}ixed-scale Restoration Unit (MRU). The RDM maps exposure features $F_{in}$ to a three-branched exposure-invariant feature $F_{n}$, while the MRU integrates the features $F^{s}$ and $F^{c}$ by the spatial-wise and channel-wise restoration, respectively.
The exposure mask predictor (EMP) assists in generating the underexposure feature $F^{u}$ and overexposure feature $F^{o}$. MaIN denotes Mask-aware IN (Instance Normalization) while Refine module represents residual channel attention block. We optimize the model with Exposure Contrastive Regularization (ECR).}
\label{fig:framework}
\end{figure*}

\section{Related Works}

\subsection{Conventional Methods}

The task of image exposure correction has been studied for a long time. Traditional methods primarily rely on manual adjustments of the model, such as histogram equalization~\cite{pizer1987adaptive, arici2009histogram,yuan2012automatic,rivera2012content}, and gamma correction~\cite{guan2009image,huang2012efficient}.
CLAHE~\cite{reza2004realization} is devised to amplify the brightness of the images with improper exposure by remapping the histograms, thereby achieving heightened contrast. 
Meanwhile, techniques grounded in the Retinex theory~\cite{land1977retinex} have emerged.
These methods decompose images into reflectance and illumination constituents.
Brightness can be improved by enhancing the illumination layer, and noise can be suppressed by regularizing the reflection layer.
For instance, LIME~\cite{guo2016lime} refines the illumination map through structured priors to achieve enhancement. RRM~\cite{li2018structure} predicts image noise while simultaneously estimating a structure-revealing reflection map and a locally smoothed illumination map.
Despite the commendable outcomes achieved by these methods in exposure correction, many of them rely on intricate manual design or grapple with excessive constraints, ultimately leading to suboptimal outcomes.

\subsection{Learning-based Methods}

Recently, deep learning-based methods~\cite{chen2018deep,yang2023implicit} have attracted much attention and greatly boosted the performance.
A notable portion of these strategies is rooted in the Retinex theory~\cite{wei2018deep,wang2019underexposed,zhang2021beyond,wu2022uretinex}, where the objective is to model the relationship between the illumination and reflectance components through extensive data-driven learning. In this domain, RetinexNet~\cite{wei2018deep} and KinD++~\cite{zhang2021beyond} introduce simple yet effective networks dedicated to reinstating these distinct components. In contrast, as another form of decomposition method, DRBN~\cite{yang2020fidelity} decomposes the features into different frequency bands and enhances them recursively.

A few recent works~\cite{afifi2021learning,nsampi2021learning,huang2022exposure,huang2022ecl,huang2023learning,wang2023decoupling} focus on intricate multiple exposure correction tasks with opposite optimization goals. For instance, MSEC~\cite{afifi2021learning} proposes a coarse-to-fine network architecture to achieve color and detail enhancements. ENC~\cite{huang2022exposure} proposes a plug-and-play module including instance normalization and compensation to map different exposures to normal exposures. ECLNet~\cite{huang2022ecl} uses the bilateral activation mechanism to model exposure-consistency representations. ERL~\cite{huang2023learning} correlates and constrains the relationship of corrected samples across the mini-batch to help the model optimization. LCDPNet~\cite{wang2022local} introduces a local color distribution operator based on Retinex theory for non-uniform exposure correction.


\section{Method}

In this section, we first introduce the motivation and overview of our proposed method. Then, the region-aware de-exposure module and mixed-scale restoration unit are described in detail. Finally, we present the exposure contrastive regularization strategy and utilized loss function.

\subsection{Motivation and Overview}

When tackling a single image with mixed exposures that incorporates large discrepancies of both overexposed and underexposed regions, it is difficult for the network to converge stably, resulting in imbalanced performance across different exposures (see Figure~\ref{fig:motivation}).
To this end, we consider locality with different exposures to reduce the adverse effects of inconsistent optimization. To achieve this, we adopt the idea of the divide and conquer strategy, and design a region-aware exposure correction framework consisting of two well-designed modules concatenated in a chain of consecutive RMBs (see Figure~\ref{fig:framework}).

More concretely, given a mixed exposure image $I_{in} \in \mathbb{R}^{H\times W \times 3}$, where $H\times W$ represents the spatial resolution of the input image, we build RECNet that stacks $N_{i \in [1, 2, 3, 4, 5]}$ RMBs to learn exposure consistency representations sequentially, followed by a residual channel attention block~\cite{hu2018squeeze} for refining the aggregated features. In addition, we propose a learnable exposure mask predictor to partition over-/under-exposed regions along the spatial dimension, which is fed to provide the localization information of exposure. For implementing the ECR strategy, we perform the constraint between the original input/output representations and the ground-truth representation.

\subsection{De-exposure and Restoration}
As shown in Figure~\ref{fig:framework}, we propose the framework containing a series of Blocks (RMB) with two components: Region-aware De-exposure Module (RDM) is mapping different local-wise exposure features to an exposure-invariant feature space, while Mixed-scale Restoration Unit (MRU) aims at integrating mix-scale features in a spatial- and channel-wise manner to recover the discriminative information reduced by the Instance Normalization (IN)~\cite{ulyanov2016instance, huang2017arbitrary} in RDM.

\subsubsection{Region-aware De-exposure Module} 
The original image $I_{in}$ is translated to an initial $\mathcal{F}$ by a $1\times1$ conv layer. In the de-exposure module, different from directly employing IN to coarsely global features~\cite{huang2022exposure}, we leverage a learnable predictor $\Theta$ consisting of a series of conv-relu blocks and ending with a sigmoid function (see supplementary material for details) to finely align respective exposure feature regions. $\Theta$ maps $\mathcal{F}$ to an underexposure mask $\mathcal{M}^u$ aligned with the spatial size of $I_{in}$.
Then the $\mathcal{M}^u$ maintains the underexposed region while simultaneously masking the complementary overexposed region in the normalization branch, and vice versa. The process is formulated as:
\begin{equation}
\begin{split}
&\mathcal{F}^{o/u} = \mathcal{F}_{in} \otimes \mathcal{M}^{o/u},~ \emph{s.t.}~\mathcal{M}^{o} \oplus \mathcal{M}^{u} = I,
\end{split}
\end{equation}
where $\mathcal{M}^{o/u}$ denotes the spatial over-/under-exposure mask, while $\mathcal{F}^o$ and $\mathcal{F}^u$ represent the overexposure part and the underexposure part, respectively. $\otimes/\oplus$ are element-wise multiplication and addition.
To prevent the generated mask from being too coarse to extract regional features, we introduce mask-aware instance normalization (MaIN) to adaptively adjust the segmented features $\mathcal{F}^{o/u}$ in a soft way. The process is formulated as:
\begin{equation}
\begin{split}
& \mathcal{F}^{o'/u'} = \sigma(f_{3 \times 3}([\mathcal{F}^{o/u}_m, \mathcal{F}^{o/u}_a, \mathcal{M}^{o/u}])) \otimes \mathcal{F}^{o/u}, \\
& \mathcal{F}^{n} = f_{1\times1}(\textrm{IN}([\mathcal{F}^{o'}, \mathcal{F}_{in}]) \oplus f_{1\times1}(\textrm{IN}([\mathcal{F}^{u'}, \mathcal{F}_{in}])
\end{split}
\end{equation}
where $\sigma(\cdot)$ denotes sigmoid function, $f_{n \times n}(\cdot)$ denotes $n\times n$ convolution, $\mathcal{F}_m$ and $\mathcal{F}_a$ represent max-pooled and average-pooled features of $\mathcal{F}^{o/u}$, and $[\cdot]$ denotes channel-wise concatenation, respectively.

\subsubsection{Mixed-scale Restoration Unit} 
Applying normalization to the features in the de-exposure module usually leads to the loss of discriminative information~\cite{choi2021meta, ren2023crossing}, especially the details of local instances.
In fact, rich multi-scale local representation has demonstrated the effectiveness~\cite{wang2020dcsfn,jiang2020multi} in better removing rain streaks. Therefore, we design the MRU by first inserting multi-scale depth-wise convolution to enrich local information. Furthermore, we introduce dual channel-wise self-attention to enhance the discriminative learning ability of the network by focusing on the information relationships between initial and normalized features. It guides the information restoration from the initial features, making the cooperative representation more efficient.
Specifically, the normalized tensor $\mathcal{F}^{n} \in \mathbb{R}^{H\times W \times C}$ is fed into two parallel branches. The 3$\times$3 and 5$\times$5 depth-wise convolutions are employed to enhance the multi-scale local information extraction. In this way, the feature transformation flows are formulated as:
\begin{equation}
\begin{split}
& \mathcal{X}_{kv}^1 = \zeta(f_{3\times 3}(\mathcal{F}^n)), \mathcal{X}_{kv}^2=\zeta(f_{5\times 5}(\mathcal{F}^n)),\\
& \mathcal{X}_k = \zeta(f_{3\times 3}([\mathcal{X}_k^1, \mathcal{X}_k^2])), \mathcal{X}_v = \zeta(f_{5\times 5}([\mathcal{X}_v^1, \mathcal{X}_v^2])), \\
& \mathcal{F}^s = f_{1\times 1}\left([\mathcal{X}_k, \mathcal{X}_v] \right), \\
\end{split}
\end{equation}
where $\mathcal{X}_{kv}^i$ denotes combined with $\mathcal{X}_k^i$ and $\mathcal{X}_v^i$, and $\zeta$ represents ReLU activation. Meanwhile, dual channel-wise self-attention (CSA) is utilized to recover details by treating initial/normalized input features as paired \textit{query/key-value} to attain attentioned $\mathcal{F}^c$. Finally, we obtain the fused feature $\mathcal{F}_{out}$ from the hybrid feature transformation to ensure the completeness of the information. The formulations are:
\begin{equation}
\begin{split}
& \mathcal{X}_{q}^1 = \zeta(f_{3\times 3}(\mathcal{F}_{in})), \mathcal{X}_{q}^2=\zeta(f_{5\times 5}(\mathcal{F}_{in})),\\
& \mathcal{A}^i = \textrm{Attn}(\mathcal{X}_{q}^i, \mathcal{X}_{k}^i, \mathcal{X}_{v}^i) = \textrm{softmax}(\frac{{\mathcal{X}_q^i}^T\mathcal{X}_{k}^i}{\lambda})\mathcal{X}_{v}^i, i = 1, 2,\\
& \mathcal{F}^c = f_{1\times 1}([\mathcal{A}^1, \mathcal{A}^2]), \mathcal{F}_{out} = f_{1\times 1}([\mathcal{F}^n, \mathcal{F}^s, \mathcal{F}^c]), \\
\end{split}
\end{equation}
where $\lambda$ is a temperature factor defined by $\lambda = \sqrt{d}$. Multi-head attention is implemented to each $k$ new $\mathcal{X}_{q}, \mathcal{X}_{k}, \mathcal{X}_{v}$, and $d=C/k$. 

\begin{figure*}[ht]
\centering
\includegraphics[width=\textwidth]{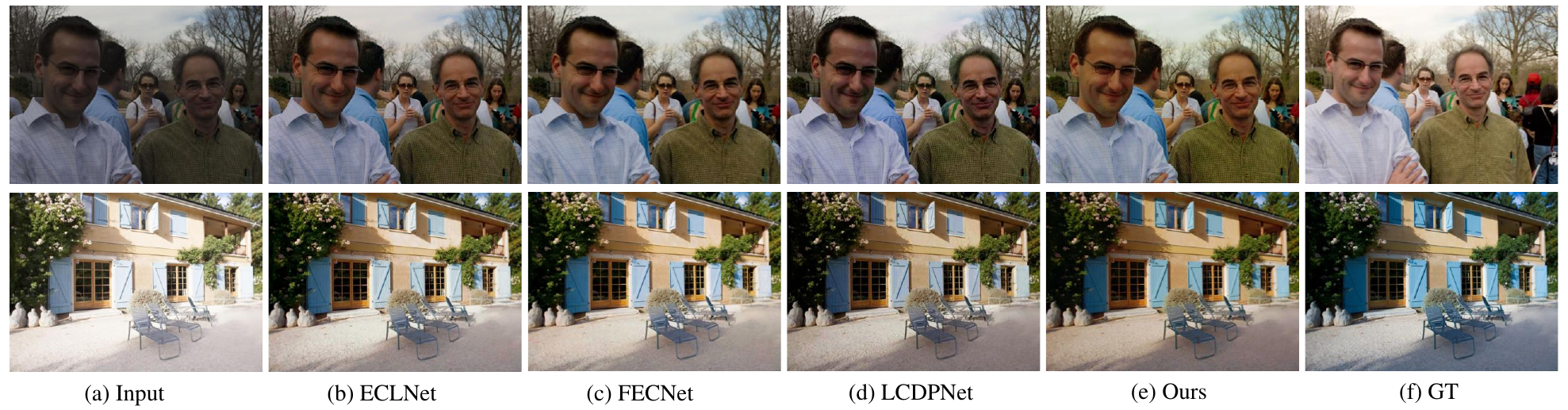}
\caption{Visualization comparable results on the MSEC dataset of (top) underexposure correction and (bottom) overexposure correction.}
\label{fig:msec}
\end{figure*}

\begin{figure*}[ht]
\centering
\includegraphics[width=\textwidth]{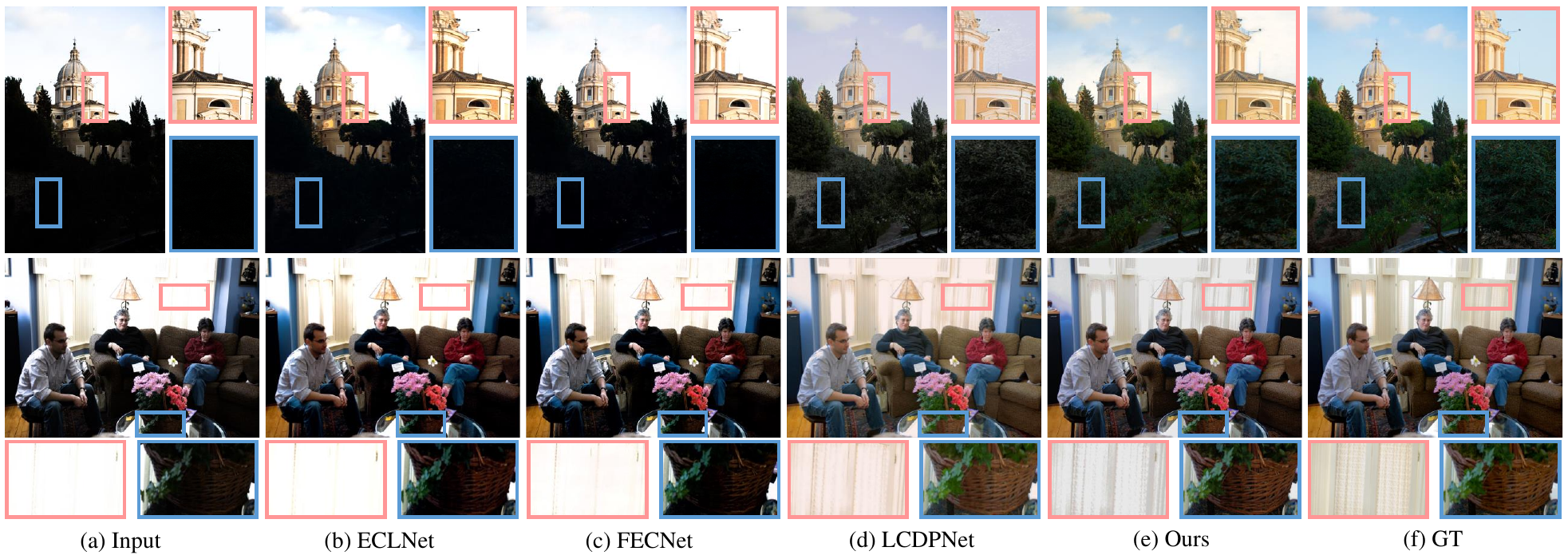}
\caption{Visualization results on the LCDP dataset of mixed exposure correction. Our model reconstructs the details in the overexposed regions (building and curtain) as well as the underexposed regions (grass and basket).}
\label{fig:lcdp}
\end{figure*}

\subsection{Exposure Contrastive Regularization Strategy}
To further promote self-exposure consistency within regions and exposure continuity across regions, we introduce the Exposure Contrastive Regularization (ECR) strategy.
The exposure distribution is often spatially non-uniform between overexposed and underexposed regions. Unlike ECLNet~\cite{huang2022exposure} that augments the local region exposures using a mask with random positions and fixed sizes, we use the mask $\mathcal{M}^u$ produced by the trainable Exposure Mask Predictor $\Theta$ to separate the overexposed and underexposed regions. The process can be formulated as:
\begin{equation}
\begin{split}
& \Omega^u = I_{out} \otimes \mathcal{M}^u, \Omega^o = I_{out} \otimes (1-\mathcal{M}^u),
\end{split}
\end{equation}
where $I_{out}$ denotes the generated image. To force the self-exposure consistency within $\Omega^o$ and $\Omega^u$ and exposure continuity across them, we also regard each kind of exposure as a style like~\cite{huang2022exposure}, and propose a style-based contrastive learning strategy to guarantee a uniform exposure distribution.
Specifically, we perform the constraint by pulling closer the distance between the corresponding regions of the generated image and the ground-truth image in the feature space, and simultaneously modeling and comparing the exposure distance between the regions in an image.
The regularization function is as follows:
\begin{equation}
\begin{split}
\mathcal{L}_{ecr} &= \frac{\mathcal{D}(h^{o/u}, {h^+}^{o/u})}{\mathcal{D}(h^{o/u}, {h^+}^{o/u}) +\mathcal{D}(h^{o/u}, {h^-}^{o/u})} \\
& + \frac{\mathcal{D}(c, c^+)}{\mathcal{D}(c, c^+) + \mathcal{D}(c, c^-)},
\end{split}
\end{equation}
where the function $\mathcal{D}(\cdot)$ is the $\mathcal{L}_1$ distance function. $h$, $h^+$, and $h^-$ are the feature vectors extracted from a pre-trained VGG16~\cite{simonyan2014very} network. The inputs are the overexposed region $\Omega^o$ or underexposed region $\Omega^u$ of the generated image $I_{out}$, ground truth $I_{gt}$, and input image $I_{in}$. The parameter $c$ represents the over-under exposure style correlation calculated by $c=\textrm{Gram}(h^o, h^u)$, $c^+=\textrm{Gram}({h^+}^o, {h^+}^u)$, and $c^-=\textrm{Gram}({h^-}^o, {h^-}^u)$.

\subsubsection{Loss Function}
We incorporate a range of loss terms to facilitate the training of our model. We employ the commonly utilized MSE term, denoted as $\mathcal{L}_{mse}$, to measure the accuracy of intensity reconstruction. In order to rectify color discrepancies, we introduce the cosine similarity term, labeled as $\mathcal{L}_{cos}$~\cite{wang2019underexposed,xu2021intensity}. This term quantifies the color similarity between the reconstructed image and its corresponding ground truth within the sRGB color space. In addition, in order to guide the network to estimate the exposure mask, we apply the binary cross entropy term $\mathcal{L}_{bce}$, which measures the residual brightness in the YCbCr space. The ground truth mask is calculated by:
\begin{eqnarray}
    \mathcal{M}^i_{gt}=\left\{
    \begin{aligned}
        1, &  & if \quad Y^i_{in}-Y^i_{gt} > 0, \\
        0, &  & otherwise,
    \end{aligned}
    \right.
\end{eqnarray}
where $Y^i$ represents the $i$-th pixel value of the Y channel of the image in the YCbCr space. We also propose an exposure contrastive regularization $\mathcal{L}_{ecr}$ constraint to further ensure exposure consistency and continuity. The overall function can be written as:
\begin{equation}
\begin{split}
\mathcal{L}_{total} &= \lambda_1\mathcal{L}_{mse} + \lambda_2\mathcal{L}_{cos} + \lambda_3\mathcal{L}_{bce} + \lambda_4\mathcal{L}_{ecr},
\end{split}
\end{equation}
where $\lambda_1$, $\lambda_2$, $\lambda_3$, and $\lambda_4$ are weight hyperparameters.


\section{Experiments}

\begin{table*}[!t]
\renewcommand{\arraystretch}{1.15}
\setlength{\tabcolsep}{0.4mm}
\begin{tabular}{ccccccccccccccc}
\hline
\multicolumn{1}{c|}{}  & \multicolumn{6}{c|}{MSEC}  & \multicolumn{6}{c|}{SICE}   & \multicolumn{2}{c}{LCDP}  \\ 
\cline{2-15}
\multicolumn{1}{c|}{Method}      & \multicolumn{2}{c|}{Under}          & \multicolumn{2}{c|}{Over}           & \multicolumn{2}{c|}{Average}        & \multicolumn{2}{c|}{Under}          & \multicolumn{2}{c|}{Over}           & \multicolumn{2}{c|}{Average} & \multicolumn{2}{c}{Average} \\ \cline{2-15}
\multicolumn{1}{c|}{}            & PSNR  & \multicolumn{1}{c|}{SSIM}   & PSNR  & \multicolumn{1}{c|}{SSIM}   & PSNR  & \multicolumn{1}{c|}{SSIM}   & PSNR  & \multicolumn{1}{c|}{SSIM}   & PSNR  & \multicolumn{1}{c|}{SSIM}   & PSNR  & \multicolumn{1}{c|}{SSIM} & PSNR  & \multicolumn{1}{c}{SSIM}   \\ \hline
\multicolumn{1}{c|}{RetinexNet (BMVC18)}  & 12.13 & \multicolumn{1}{c|}{0.6209} & 10.47 & \multicolumn{1}{c|}{0.5953} & 11.14 & \multicolumn{1}{c|}{0.6048} & 12.94 & \multicolumn{1}{c|}{0.5171} & 12.87 & \multicolumn{1}{c|}{0.5252} & 12.90 & \multicolumn{1}{c|}{0.5212} & 19.25 & \multicolumn{1}{c}{0.7041}    \\
\multicolumn{1}{c|}{RUAS (CVPR21)}        & 13.43 & \multicolumn{1}{c|}{0.6807} & 6.39 & \multicolumn{1}{c|}{0.4655} & 9.20 & \multicolumn{1}{c|}{0.5515} & 16.63 & \multicolumn{1}{c|}{0.5589} & 4.54  & \multicolumn{1}{c|}{0.3196} & 10.59 & \multicolumn{1}{c|}{0.4393}  & 13.76 & \multicolumn{1}{c}{0.6060} \\
\multicolumn{1}{c|}{SCI (CVPR22)}         & 9.97  & \multicolumn{1}{c|}{0.6681} & 5.837 & \multicolumn{1}{c|}{0.5190} & 7.49  & \multicolumn{1}{c|}{0.5786} & 17.86 & \multicolumn{1}{c|}{0.6401} & 4.45  & \multicolumn{1}{c|}{0.3629} & 12.49 & \multicolumn{1}{c|}{0.5051} & 11.87 & \multicolumn{1}{c}{0.5234}\\
\multicolumn{1}{c|}{LCDPNet (ECCV22)}     & 22.35 & \multicolumn{1}{c|}{\underline{0.8650}} & 22.17 & \multicolumn{1}{c|}{0.8476} & 22.30 & \multicolumn{1}{c|}{0.8552} & 17.45 & \multicolumn{1}{c|}{0.5622} & 17.04 & \multicolumn{1}{c|}{0.6463} & 17.25 & \multicolumn{1}{c|}{0.6043} & \underline{23.24} & \multicolumn{1}{c}{\underline{0.8420}} \\
\multicolumn{1}{c|}{FECNet (ECCV22)}      & \underline{22.96} & \multicolumn{1}{c|}{0.8598} & \underline{23.22} & \multicolumn{1}{c|}{0.8748} & \underline{23.12} & \multicolumn{1}{c|}{\underline{0.8688}} & 22.01 & \multicolumn{1}{c|}{0.6737} & \underline{19.91} & \multicolumn{1}{c|}{0.6961} & \underline{20.96} & \multicolumn{1}{c|}{0.6849}  & 22.41 & \multicolumn{1}{c}{0.8402}\\
\multicolumn{1}{c|}{DRBN-ENC (CVPR22)}    & 22.72 & \multicolumn{1}{c|}{0.8544} & 22.11 & \multicolumn{1}{c|}{0.8521} & 22.35 & \multicolumn{1}{c|}{0.8530} & 21.89 & \multicolumn{1}{c|}{0.7071} & 19.09 & \multicolumn{1}{c|}{\underline{0.7229}} & 20.49 & \multicolumn{1}{c|}{\underline{0.7150}} & 22.09 & \multicolumn{1}{c}{0.8271}\\
\multicolumn{1}{c|}{MSEC (CVPR21)}        & 20.52 & \multicolumn{1}{c|}{0.8129} & 19.79 & \multicolumn{1}{c|}{0.8156} & 20.08 & \multicolumn{1}{c|}{0.8210} & 19.62 & \multicolumn{1}{c|}{0.6512} & 17.59 & \multicolumn{1}{c|}{0.6560} & 18.58 & \multicolumn{1}{c|}{0.6536} & 20.38 & \multicolumn{1}{c}{0.7800}\\
\multicolumn{1}{c|}{MSEC+DA (CVPR23)}        & 21.53 & \multicolumn{1}{c|}{0.8590} & 21.55 & \multicolumn{1}{c|}{\underline{0.8750}} & 21.54 & \multicolumn{1}{c|}{0.8670} & 20.94 & \multicolumn{1}{c|}{\textbf{0.7546}} & 17.49 & \multicolumn{1}{c|}{0.6640} & 19.22 & \multicolumn{1}{c|}{0.7093}  & 21.05 & \multicolumn{1}{c}{0.8119}\\
\multicolumn{1}{c|}{ECLNet (ACMMM22)}      & 22.37 & \multicolumn{1}{c|}{0.8566} & 22.70 & \multicolumn{1}{c|}{0.8673} & 22.57 & \multicolumn{1}{c|}{0.8631} & 22.05 & \multicolumn{1}{c|}{0.6893} & 19.25 & \multicolumn{1}{c|}{0.6872} & 20.65 & \multicolumn{1}{c|}{0.6861} & 22.44 & \multicolumn{1}{c}{0.8061}\\
\multicolumn{1}{c|}{ECLNet+ERL (CVPR23)}  & 22.90 & \multicolumn{1}{c|}{0.8624} & 22.58 & \multicolumn{1}{c|}{0.8676} & 22.70 & \multicolumn{1}{c|}{0.8655} & \underline{22.14} & \multicolumn{1}{c|}{0.6908} & 19.47 & \multicolumn{1}{c|}{0.6982} & 20.81 & \multicolumn{1}{c|}{0.6945}  & 22.63 & \multicolumn{1}{c}{0.8096}\\ \hline
\multicolumn{1}{c|}{RECNet (Ours)}   
&  \textbf{23.57}  & \multicolumn{1}{c|}{\textbf{0.8655}} & \textbf{23.81}  & \multicolumn{1}{c|}{\textbf{0.8765}} & \textbf{23.69}  & \multicolumn{1}{c|}{\textbf{0.8699}} & \textbf{22.50} & \multicolumn{1}{c|}{\underline{0.7254}} & \textbf{20.81} & \multicolumn{1}{c|}{\textbf{0.7363}} & \textbf{21.66} & \multicolumn{1}{c|}{\textbf{0.7309}}  & \textbf{23.54} & \multicolumn{1}{c}{\textbf{0.8468}}\\
\hline
\end{tabular}
\caption{Quantitative results of different methods on the MSEC and the SICE and LCDP datasets in terms of PSNR and SSIM. A higher metric indicates better performance. The best and second results are marked in bold and underlined, respectively. Notably, LCDP dataset only contains mixed exposure images, so we do not classify over- or under-exposures.}
\label{table:baseline}
\end{table*}

\subsection{Experimental Settings}

\subsubsection{Datasets}
Our network is trained on two representative multiple exposure datasets, including the Multi-Scale Exposure Correction (MSEC) dataset~\cite{afifi2021learning} and the Single Image Contrast Enhancement (SICE) dataset~\cite{cai2018learning}. The MSEC dataset is rendered from the MIT-Adobe FiveK dataset~\cite{bychkovsky2011learning}, consisting of five exposure levels and corrected by experts as the ground truth. As for the SICE dataset, we set the same division settings for the training/validation/test subsets as~\cite{huang2023learning} for the above two datasets.
As existing datasets contain mostly images with either overexposure or underexposure, we select the paired dataset produced by LCDPNet~\cite{wang2022local}, consisting of 1,733 pairs of images, which are split into 1,415 for training, 100 for validation, and 218 for testing. It is also rendered from MIT-Adobe FiveK raw images and the exposure levels. The images contain non-uniformly illuminated scenes, including both overexposure and underexposure regions, which are adjusted with a specific linear transformation function.

\subsubsection{Implementation Details}
We conduct our experiments on a single NVIDIA GeForce RTX3090. Our proposed model is implemented using Pytorch. The parameters of the network are optimized by the ADAM optimizer with $\beta_1=0.9$ and $\beta_2=0.99$. The learning rate is 0.0001, while the batch size is set to 8. The images apply random horizontal and vertical flipping to augment the input data. The weights for the terms in the loss function in Eq. (8) are $\lambda_1=\lambda_2$ = 1.0, $\lambda_3$ = 0.25, and $\lambda_4$ = 0.1.

\begin{figure}[!t]
\centering
\includegraphics[width=\columnwidth]{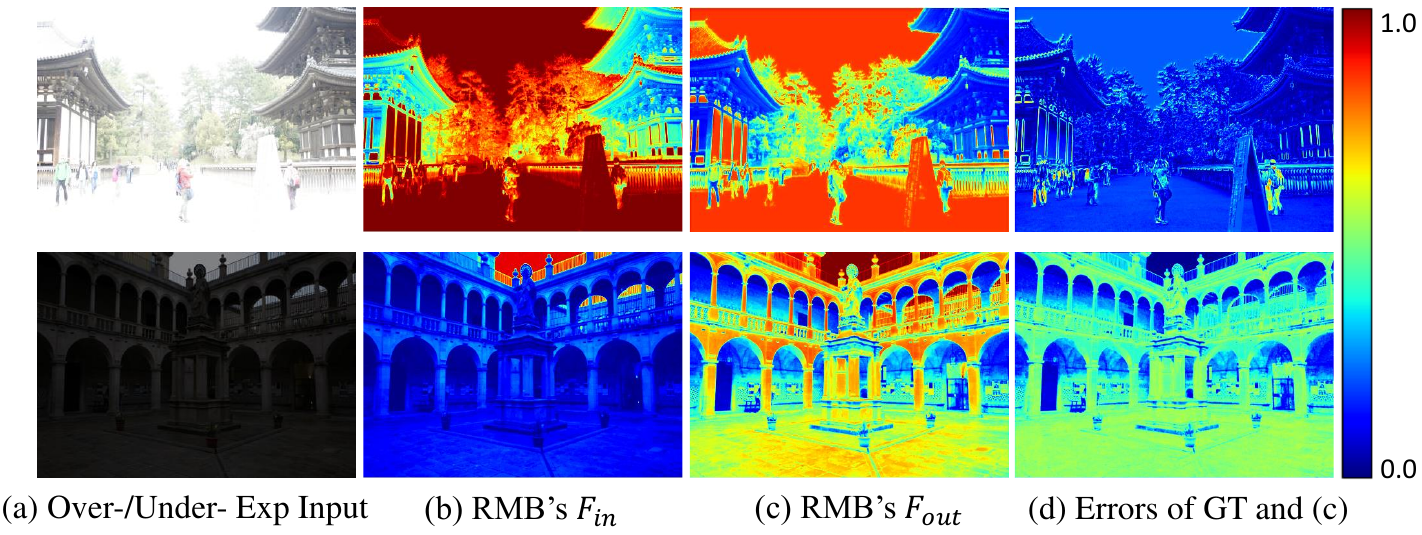}
\caption{The visualization of the input/output feature of RMB module.}
\label{fig:feat_vis}
\end{figure}

\subsection{Comparison with State-of-the-art Methods}
To verify the effectiveness of our method, several exposure correction and low-light image enhancement methods are adopted as a comparison. We select both conventional enhancement methods and deep-learning-based methods, including RetinexNet~\cite{wei2018deep}, RUAS~\cite{liu2021retinex}, SCI~\cite{ma2022toward}, LCDPNet~\cite{wang2022local}, FECNet~\cite{huang2022deep}, DRBN~\cite{yang2020fidelity}, MSEC~\cite{afifi2021learning}, and ECLNet~\cite{huang2022ecl}.
Decoupling-and-Aggregating~\cite{wang2023decoupling}, ENC~\cite{huang2022exposure}, and ERL~\cite{huang2023learning} are effective play-and-plugin methods. We adopt the commonly used Peak Signal-to-Noise Ratio (PSNR) and Structural Similarity (SSIM) as our evaluation metrics.

\subsubsection{Quantitative Evaluation}
We present the quantitative comparison results in Table~\ref{table:baseline}. As we can see, our method outperforms other methods on the MSEC~\cite{afifi2021learning} and SICE~\cite{cai2018learning} datasets in terms of both underexposure and overexposure cases. On the LCDP~\cite{wang2022local} dataset with mixed exposure, our method has a marginal improvement in quantitative performance compared to LCDPNet.
From the results, our RECNet achieves the highest score for both PSNR and SSIM.
Furthermore, our model is lightweight, illustrating its potential to be applied to computationally compact devices.

\subsubsection{Qualitative Evaluation}
\begin{figure}[!t]
\centering
\includegraphics[width=\columnwidth]{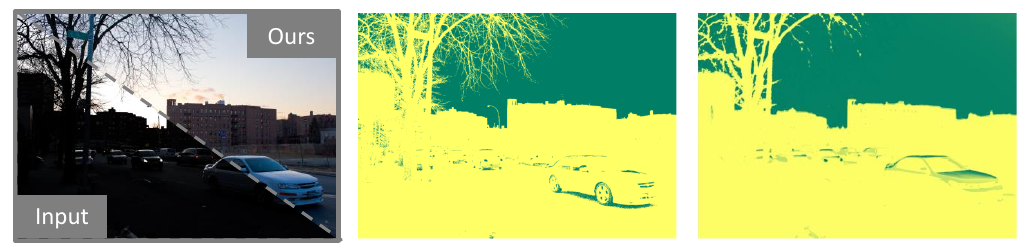}
\caption{The visualization of the masks of the ground truth (middle) and the prediction from the EMP module (right).}
\label{fig:mask}
\end{figure}

We provide visual comparisons between our method and those existing baseline methods, including ECLNet~\cite{huang2022ecl}, FECNet~\cite{huang2022deep}, and LCDPNet~\cite{wang2022local}. Figure~\ref{fig:msec} shows the correction effects on the MSEC dataset. It can be seen that our algorithm also achieves the best pleasing effect in terms of lightness and color saturation.
We further visually compare the results on the LCDP dataset~\cite{wang2022local} in Figure~\ref{fig:lcdp}. As can be seen, our method can correct both overexposed and underexposed regions and produce better color and structure recovery effects.
As shown in the last column of Figure~\ref{fig:feat_vis}, the errors between underexposure and overexposure features are smaller after being processed by our method.

\begin{figure}[ht]
\centering
\includegraphics[width=0.9\columnwidth]{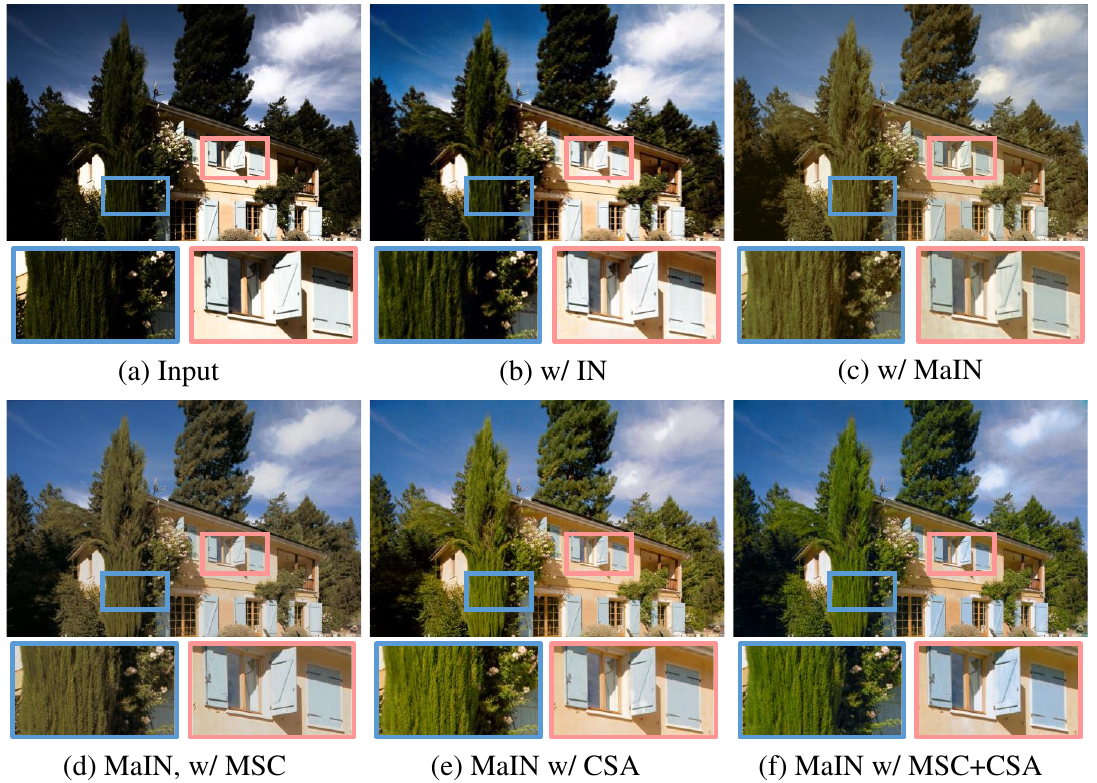}
\caption{The visualization of the ablation study. MaIN represents the Mask-aware Instance Normalization. MDC and CSA denote multi-scale convolution and channel-wise self-attention, respectively.}
\label{fig:ablation}
\end{figure}

\begin{figure}[ht]
\centering
\includegraphics[width=0.9\columnwidth]{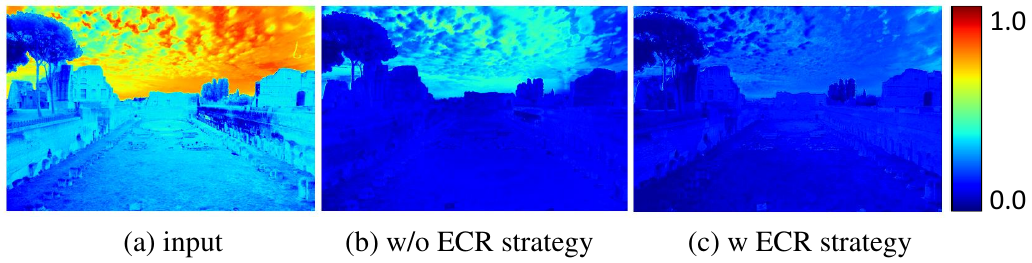}
\caption{Visualization of the errors between the exposure-corrected feature and the corresponding ground truth feature. The errors are further reduced with $\mathcal{L}_{ecr}$. The result maintains consistency across different exposure regions.}
\label{fig:lecr}
\end{figure}

\subsection{Ablation Studies and Discussion}

In this section, we conduct ablation studies to demonstrate the effectiveness of our proposed method, including the design of the EMP and RMB modules, the ECR strategy, and the RECNet.

\subsubsection{Effectiveness of the specific modules}

With the constraint of the multi-branch exposure map, the model learns the region-aware exposure correction adaptively, as shown in Figure~\ref{fig:mask}. 
We also conduct several experiments to demonstrate the reasonableness of the RMB module design on the LCDP~\cite{wang2022local} dataset.
Specifically, as presented in Table~\ref{table:ablation} and Figure~\ref{fig:ablation}, with the equipment of Mask-aware IN operation, the problems of overexposure and underexposure are significantly improved compared with IN. Additionally, the adoption of the multi-scale convolution and channel-wise self-attention schemes in MRU obtains improvements, demonstrating their effectiveness in capturing discriminative information such as the color saturation of the local instances. By adding exposure consistency regularization scheme, the model can further balance the exposure intensity of different regions and continuity (see Figure~\ref{fig:lecr}).

\subsubsection{Investigation of the number of parameters in RECNet}

\begin{table}[!t]
\renewcommand{\arraystretch}{1.05}
\setlength{\tabcolsep}{1.5mm}
\begin{tabular}{c|ccccc|cc}
\hline
\multirow{2}{*}{Model} & \multicolumn{2}{c|}{RDM} & \multicolumn{2}{c|}{MRU} & \multirow{2}{*}{$\mathcal{L}_{ecr}$} & \multirow{2}{*}{PSNR} & \multirow{2}{*}{SSIM} \\ \cline{2-5}
 & IN & \multicolumn{1}{c|}{MaIN} & MSC & \multicolumn{1}{c|}{CSA} &          &                       &                       \\ \hline
(a)   &  \checkmark  &      &     &     &     &   22.09  &   0.8292   \\
(b)   &    &  \checkmark    &     &     &     &   22.41   &  0.8305    \\ 
(c)   &    &  \checkmark    &  \checkmark   &     &     &  22.98    &  0.8386    \\
(d)   &    &    \checkmark  &     &   \checkmark  &     &   23.14   &  0.8394   \\
(e)   &    &   \checkmark   &   \checkmark  &   \checkmark  &    & 23.37     &   0.8431   \\
(f)   &    &    \checkmark  &  \checkmark   &   \checkmark  &  \checkmark   &  23.54    &  0.8468    \\   \hline
\end{tabular}
\caption{Ablation study for investigating the components of the specific modules. MaIN represents the Mask-aware Instance Normalization. MDC and CSA denote multi-scale convolution and channel-wise self-attention, respectively. $\mathcal{L}_{ecr}$ is the exposure contrastive regularization loss.}
\label{table:ablation}
\end{table}

\begin{figure}[ht]
\centering
\subfloat[]{\includegraphics[width=0.49\columnwidth]{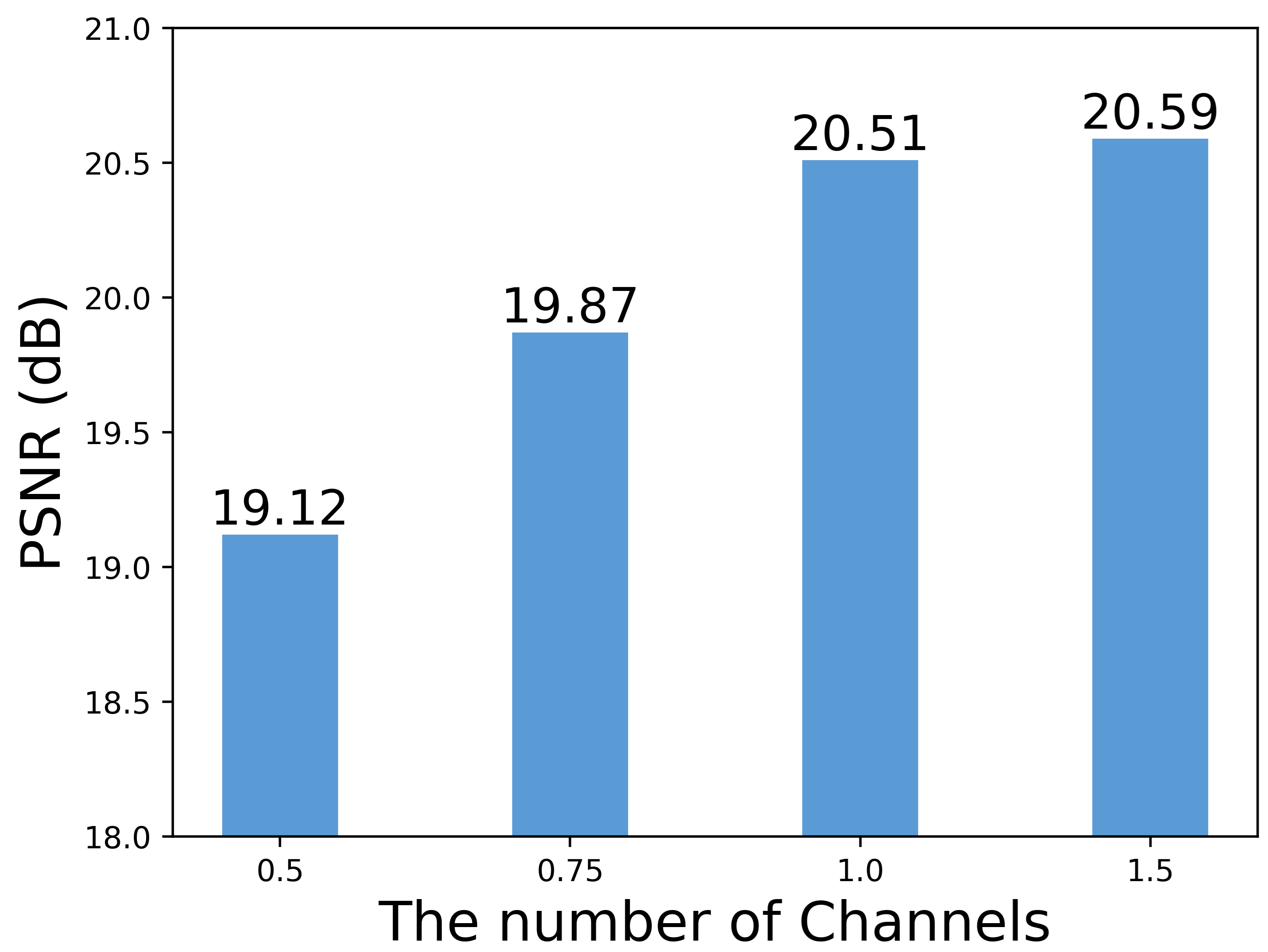}}
\subfloat[]{\includegraphics[width=0.49\columnwidth]{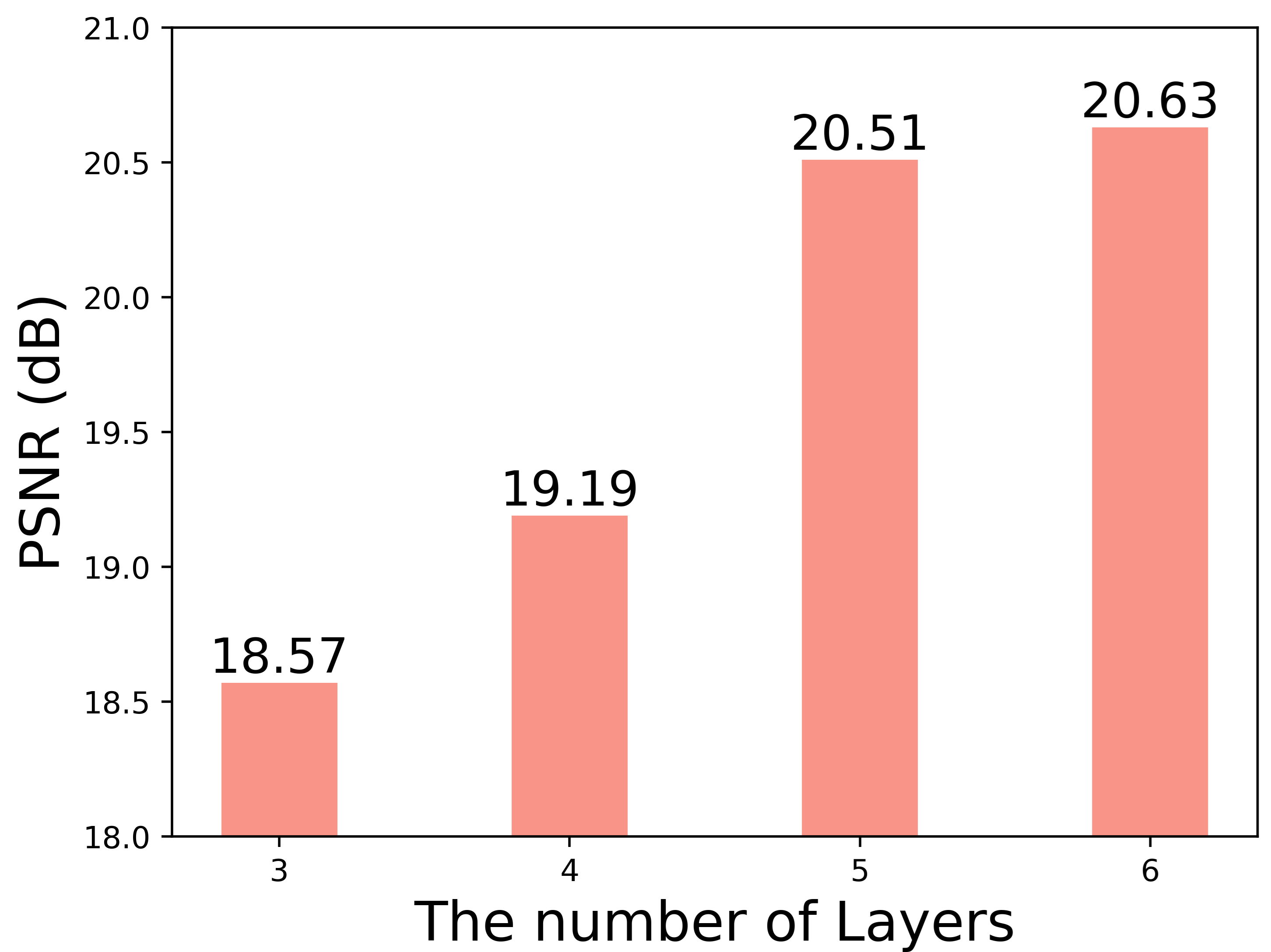}}
\caption{The ablation studies of the number of parameters in RECNet.}
\label{fig:layer_channel}
\end{figure}

We first investigate by adjusting the number of channels in each layer to 0.5, 0.75, and 1.5 times the original setting. We set the number of layers to 5 by default.
Additionally, we perform ablation studies to investigate the number of RMB modules. As shown in Figure~\ref{fig:layer_channel}, the essence results on SICE~\cite{cai2018learning} dataset demonstrate the effectiveness of the proposed network.

\subsubsection{Extension} 

To further explore the potentials of the region-aware de-exposure module (RDM) in our method, we plug it into several baselines, including DRBN~\cite{yang2020fidelity} and SID~\cite{chen2018learning}.
Under the optimization of the RDM, the performance of the baselines is improved by more than 2 dB PSNR and 0.02 SSIM on average.
The quantitative and qualitative results are presented in supplementary material. With the help of region awareness, DRBN-RDM and SID-RDM achieve better performance, which confirms the effectiveness and superiority of our proposed method.


\section{Conclusion}

In this paper, we introduce a region-aware exposure correction network for mixed exposure correction.
Specifically, we introduce an effective region-aware de-exposure module to separate the exposure-related regions and map the regional features into an exposure-invariant feature space.
Then a mixed-scale restoration unit is proposed to integrate exposure-invariant features and unprocessed features to ensure the completeness of discriminative information. We further develop an exposure contrastive regularization strategy to achieve a uniform exposure distribution across regions. Extensive experiments show that our method performs favorably against state-of-the-art methods.

\section{Acknowledgements}

This work is supported in part by the Natural Science Foundation of China under No.62272059, the National Key R\&D Program of China under No.2023YFF0904800, the Beijing Nova Program under No.20230484406, the Innovation Research Group Project of NSFC (61921003), and the 111 Project (B18008).

\bibliography{aaai24}

\end{document}